\documentclass{llncs}

\usepackage{cite}
\usepackage[pdftex]{graphicx}
\usepackage[export]{adjustbox}

\usepackage{hyperref}

\usepackage{lineno}

\usepackage{url}
\usepackage{booktabs}
\usepackage{graphicx}
\usepackage{subfigure}
\usepackage{amsmath}
\usepackage{amssymb}
\usepackage{dblfloatfix}
\usepackage{color}
\usepackage{tabularx}
\usepackage{soul}

\newcolumntype{Y}{>{\centering\arraybackslash}X}
\newcolumntype{Z}{>{\raggedleft\arraybackslash}p{0.5\textwidth}}

\newcommand{\blue}[1]{{\color{blue} #1}}
\newcommand{\anon}[1]{\blue{anonymous}}
\renewcommand{\anon}[1]{{#1}}
\DeclareMathOperator{\softmax}{softmax}

\DeclareMathOperator{\softplus}{softplus}
\newcommand{\mybold}[1]{\boldsymbol{\mathbf{#1}}}

\newcommand{\bb}{\mybold{b}}
\newcommand{\cb}{\mybold{c}}
\newcommand{\eb}{\mybold{e}}
\newcommand{\fb}{\mybold{f}}

\newcommand{\hb}{\mybold{h}}
\newcommand{\ib}{\mybold{i}}
\newcommand{\ob}{\mybold{o}}

\newcommand{\vb}{\mybold{v}}
\newcommand{\xb}{\mybold{x}}

\newcommand{\mub}{\mybold{\mu}}

\newcommand{\Xb}{\mybold{X}}

\newcommand{\Wb}{\mybold{W}}

\newcommand{\thetab}{\mybold{\theta}}
\newcommand{\pib}{\mybold{\pi}}

\newcommand{\enc}{\text{(enc)}}
\newcommand{\dec}{\text{(dec)}}

\begin{document}

\title{Unsupervised Learning for Surgical Motion by Learning to Predict the Future}


\author{Robert DiPietro \and Gregory D. Hager}

\institute{Department of Computer Science, Johns Hopkins University, Baltimore, MD, USA}

\maketitle

\begin{abstract}
We show that it is possible to learn meaningful representations of surgical motion, without supervision, by learning to predict the future. An architecture that combines an RNN encoder-decoder and mixture density networks (MDNs) is developed to model the conditional distribution over future motion given past motion. We show that the learned encodings naturally cluster according to high-level activities, and we demonstrate the usefulness of these learned encodings in the context of information retrieval, where a database of surgical motion is searched for suturing activity using a motion-based query. Future prediction with MDNs is found to significantly outperform simpler baselines as well as the best previously-published result for this task, advancing state-of-the-art performance from an F1 score of $0.60 \pm 0.14$ to $0.77 \pm 0.05$.
\end{abstract}

\setcounter{footnote}{0}

\section{Introduction} \label{introduction}

\begin{figure}[t]
\centering
\subfigure[Encoding the Past]{\includegraphics[width=1.75in]{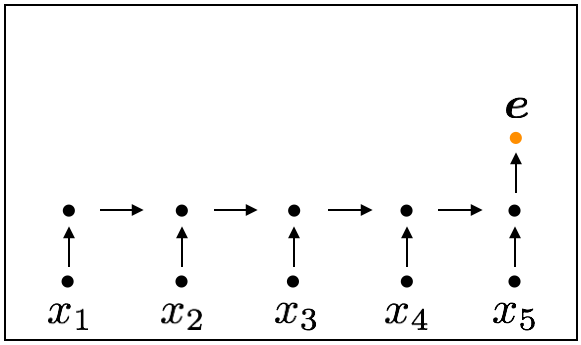}} \qquad
\subfigure[Decoding the Future]{\includegraphics[width=1.75in]{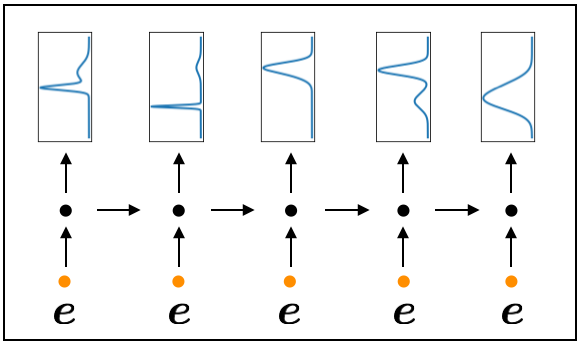}}
\caption{The encoder-decoder architecture used in this work, but using only a single kinematic signal and lengths $T_p = T_f = 5$ for visualization. More accurately, each $\xb_t \in \mathbb{R}^{n_x}$, and each time step in the future yields a multivariate mixture.}
\label{fp-diagram}
\end{figure}

Robot-assisted surgery has led to new opportunities to study human performance of surgery by enabling scalable, transparent capture of high-quality surgical-motion data in the form of surgeon hand movement and stereo surgical video. This data can be collected in simulation, benchtop training, and during live surgery, from novices in training and from experts in the operating room. This has in turn spurred new research areas such as automated skill assessment and automated feedback for trainees \cite{ahmidi2017, vedula2016, chen2016, reiley2008}.

Although the ability to capture data is practically unlimited, a key barrier to progress has been the focus on supervised learning, which requires extensive manual annotations. Unlike the surgical-motion data itself, annotations are difficult to acquire, are often subjective, and may be of variable quality. In addition, many questions surrounding annotations remain open. For example, should they be collected at the low level of gestures \cite{ahmidi2017}, at the higher level of maneuvers \cite{gao2016icra}, or at some other granularity? Do annotations transfer between surgical tasks? And how consistent are annotations among experts?

We show that it is possible to learn meaningful representations of surgery from the data itself, \emph{without the need for explicit annotations}, by searching for representations that can reliably predict future actions, and we demonstrate the usefulness of these representations in an information-retrieval setting. The most relevant prior work is \cite{gao2016ijcars}, which encodes short windows of kinematics using denoising autoencoders, and which uses these representations to search a database using motion-based queries. Other unsupervised approaches include activity alignment under the assumption of identical structure \cite{gao2016icra} and activity segmentation using hand-crafted pipelines \cite{despinoy2016}, structured probablistic models \cite{krishnan2017}, and clustering \cite{zia2017}.

Contrary to these approaches, we hypothesize that if a model is capable of predicting the future then it must encode contextually relevant information. Our approach is similar to prior work for learning video representations \cite{srivastava2015}, however unlike \cite{srivastava2015} we leverage mixture density networks and show that they are crucial to good performance. Our contributions are 1) introducing a recurrent-neural-network (RNN) encoder-decoder architecture with MDNs for predicting future motion and 2) showing that this architecture learns encodings that perform well both qualitatively (Figs. \ref{tsne} and \ref{query-vis}) and quantitatively (Table \ref{query-table}).

\section{Methods} \label{methods}

To obtain meaningful representations of surgical motion without supervision, we predict future motion from past motion. More precisely, letting $\Xb_p \equiv \{\xb_t\}_{1}^{T_p}$ denote a subsequence of kinematics from the past and $\Xb_f \equiv \{\xb_t\}_{T_p + 1}^{T_p + T_f}$ denote the kinematics that follow, we model the conditional distribution $p(\Xb_f \mid \Xb_p)$. This is accomplished with an architecture that combines an RNN encoder-decoder with mixture density networks, as illustrated in Figure \ref{fp-diagram}.

\subsection{Recurrent Neural Networks and Long Short-Term Memory}

Recurrent neural networks (RNNs) are a class of neural networks that share parameters over time, and which are naturally suited to modeling sequential data. The simplest variant is that of Elman RNNs \cite{elman1990}, but they are rarely used because they suffer from the \emph{vanishing gradient problem} \cite{bengio1994}. Long short-term memory (LSTM) \cite{hochreiter1997, gers2000fg} was introduced to alleviate the vanishing-gradient problem, and has since become one of the most widely-used RNN architectures, achieving state-of-the-art performance in many domains, including surgical activity recognition \cite{dipietro2016}. The variant of LSTM used here is
\begin{align}
\fb_t &= \sigma(\Wb_{fh} \hb_{t-1} + \Wb_{fx} \xb_{t} + \bb_f) \label{lstmbegin} &
    \ib_t &= \sigma(\Wb_{ih} \hb_{t-1} + \Wb_{ix} \xb_{t} + \bb_i) \\
\ob_t &= \sigma(\Wb_{oh} \hb_{t-1} + \Wb_{ox} \xb_{t} + \bb_o) &
    \tilde{\cb}_t &= \tanh(\Wb_{ch} \hb_{t-1} + \Wb_{cx} \xb_{t} + \bb_c) \\
\cb_t &= \fb_t \odot \cb_{t-1} + \ib_t \odot \tilde{\cb}_t &
    \hb_t &= \ob_t \odot \tanh(\cb_t) \label{lstmend}
\end{align}
where $\sigma(\cdot)$ denotes the element-wise sigmoid function and $\odot$ denotes element-wise multiplication. $\fb_t$, $\ib_t$, and $\ob_t$ are known as the forget, input, and output gates, and all weight matrices $\Wb$ and all biases $\bb$ are learned.

\subsection{The RNN Encoder-Decoder}

\begin{figure}[t]
\centering
(a) -FP MDN \quad \raisebox{-.5\height}{\includegraphics[width=3.0in]{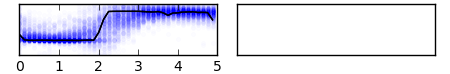}}\\
(b) FP -MDN \quad \raisebox{-.5\height}{\includegraphics[width=3.0in]{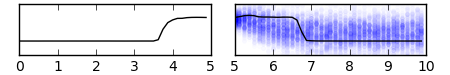}}\\
(c) FP MDN\phantom{-} \quad \raisebox{-.5\height}{\includegraphics[width=3.0in]{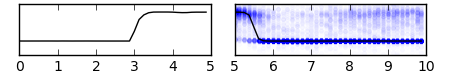}}
\caption{Visualization of predictions. Inputs and ground truth (black) are shown along with predictions (blue). -FP MDN compresses and reconstructs the past; FP -MDN predicts one blurred future; and FP MDN predicts multiple possible futures.}
\label{fp}
\end{figure}

RNN encoder-decoders \cite{cho2014} were introduced in machine translation to encode a source sentence in one language and decode it in another language, by modeling the discrete distribution $p(\text{target sentence} \mid \text{source sentence})$. We proceed similarly, by modeling the continuous conditional distribution $p(\Xb_f \mid \Xb_p)$, using LSTM for both the encoder and the decoder, as illustrated in Figure \ref{fp-diagram}.

The encoder LSTM maps $\Xb_p$ to a series of hidden states through Equations \ref{lstmbegin} to \ref{lstmend}, and the final hidden state is used as our fixed-length encoding of $\Xb_p$. Collecting the encoder's weights and biases into $\thetab^\enc$,
\begin{equation}
\eb \equiv \hb_{T_p}^\enc = f(\Xb_p ; \thetab^\enc)
\end{equation}
Similarly, the LSTM decoder, with its own parameters $\thetab^\dec$, maps $\eb$ to a series of hidden states, where hidden state $t$ is used to decode the kinematics at time step $t$ of the future. The simplest possible estimate is then $\hat{\xb}_t = \Wb \, \hb_t^\dec + \bb$, where training equates to minimizing sum-of-squares error. However, this approach corresponds to maximizing likelihood under a unimodal Gaussian, which is insufficient because distinct futures are blurred into one (see Figure \ref{fp}).

\subsection{Mixture Density Networks}

MDNs \cite{bishop1994} use neural networks to produce conditional distributions with greater flexibility. Here, we associate each time step of the future with its own mixture of multivariate Gaussians, with parameters that depend on $\Xb_p$ through the encoder and decoder. For each time step, every component $c$ is associated with a mixture coefficient $\pi_t^{(c)}$, a mean $\mub_t^{(c)}$, and a diagonal covariance matrix with entries collected in $\vb_t^{(c)}$. These parameters are computed via
\begin{align}
\pib_t(\hb_t^\dec) &= \softmax(\Wb_\pi \, \hb_t^\dec + \bb_\pi) \\
\mub_t^{(c)}(\hb_t^\dec) &= \Wb_\mu^{(c)} \, \hb_t^\dec + \bb_\mu^{(c)} \\
\vb_t^{(c)}(\hb_t^\dec) &= \softplus(\Wb_v^{(c)} \, \hb_t^\dec + \bb_v^{(c)})
\end{align}
where the $\softplus$ is used to ensure that $\vb_t^{(c)}$ has all positive elements and where the $\softmax$ is used to ensure that $\pib_t$ has positive elements that sum to 1.

We emphasize that all $\pi_t^{(c)}$, $\mub_t^{(c)}$ and $\vb_t^{(c)}$ depend implicitly on $\Xb_p$ and on the encoder's and decoder's parameters through $\hb_t^\dec$, and that the individual components of $\xb_t$ are \emph{not} conditionally independent under this model. However, in order to capture global context rather than local properties such as smoothness, we do not condition each $\xb_{t+1}$ on $\xb_t$; instead, we condition each $\xb_t$ only on $\Xb_p$ and assume independence over time steps. Our final model is then
\begin{equation}
\label{likelihood}
p(\Xb_f \mid \Xb_p) =
\prod_{\xb_t \in \Xb_f} \! \!
\sum_c
\pi_t^{(c)}(\hb_t^\dec) \,
\mathcal{N}\!\left(\xb_t \,;~ \mub_t^{(c)}(\hb_t^\dec), \vb_t^{(c)}(\hb_t^\dec)\right)
\end{equation}

\subsection{Training}

\begin{figure}[t]
\centering
\subfigure[-FP MDN]{\includegraphics[width=1.25in]{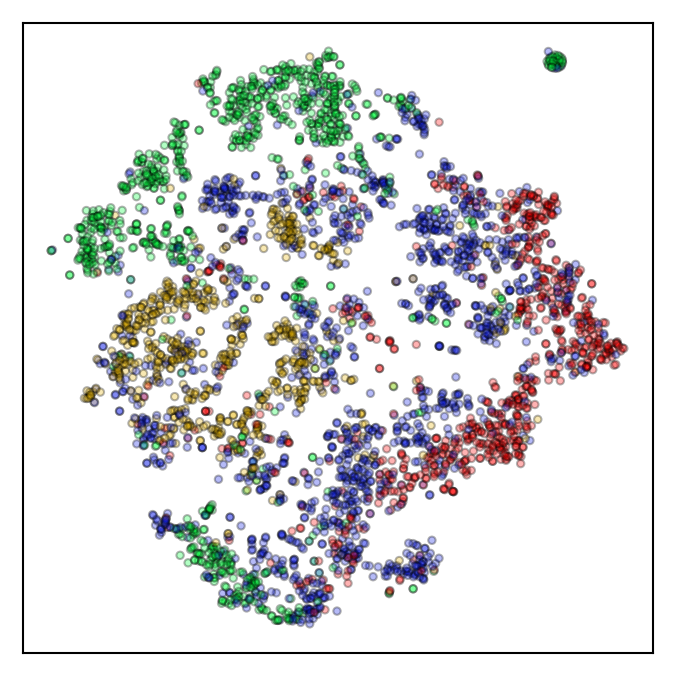}} \quad
\subfigure[FP -MDN]{\includegraphics[width=1.25in]{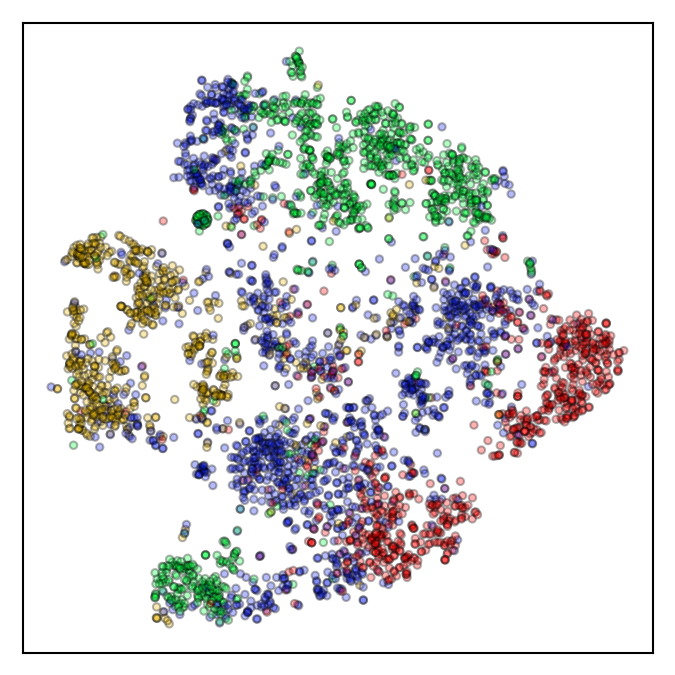}} \quad
\subfigure[FP MDN]{\includegraphics[width=1.25in]{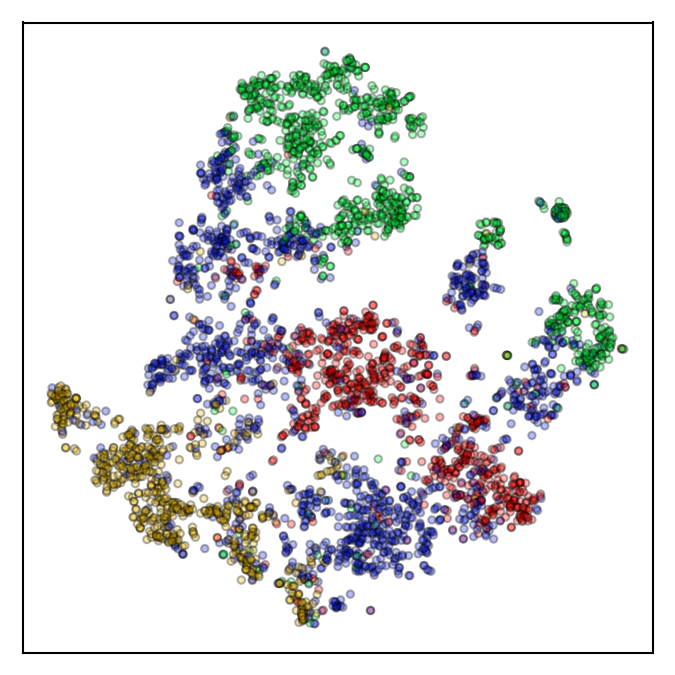}}
\caption{2-D dimensionality reductions of our 64-D encodings, obtained using t-SNE, and colored according to activity: Suture Throw (green), Knot Tying (orange), Grasp Pull Run Suture (red), and Intermaneuver Segment (blue). The activity annotations are used for visualization only. Future prediction and MDNs both lead to more separation between high-level activities in the encoding space.}
\label{tsne}
\end{figure}

\begin{figure}[h]
\centering
\subfigure[Representative -FP MDN example. Precision: 0.47. Recall: 0.78. F1 Score: 0.59.]{\includegraphics[width=5.0in]{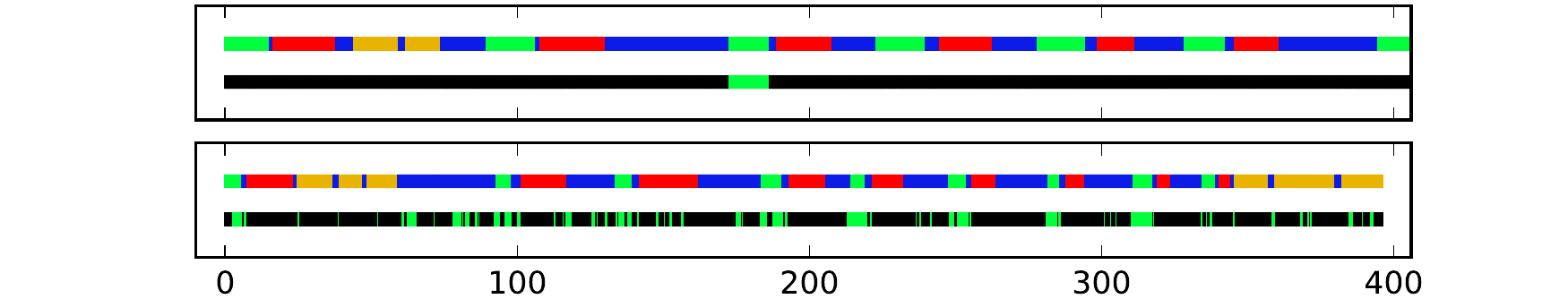}}
\subfigure[Representative FP -MDN example. Precision: 0.54. Recall: 0.70. F1 Score: 0.61.]{\includegraphics[width=5.0in]{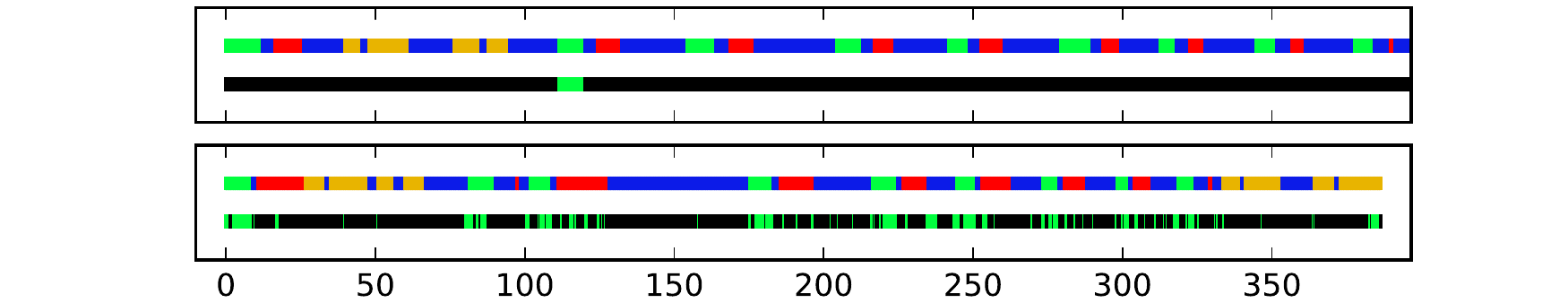}}
\subfigure[Representative FP MDN example. Precision: 0.76. Recall: 0.78. F1 Score: 0.77.]{\includegraphics[width=5.0in]{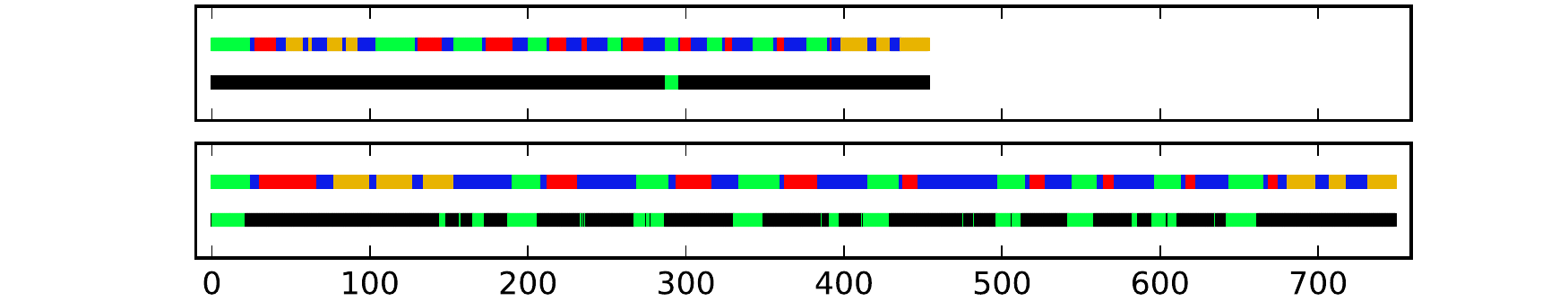}}
\caption{Qualitative results for kinematics-based suturing queries. For each example, from top to bottom, we show 1) a full activity sequence from one subject; 2) the segment used as a query; 3) a full activity sequence from a \emph{different} subject; and 4) the retrieved frames from our query. These examples were chosen because they exhibit precisions, recalls, and F1 scores that are close to the averages reported in Table \ref{query-table}.}
\label{query-vis}
\end{figure}

Given past, future pairs $(\Xb_p^{(n)}, \Xb_f^{(n)})$, training is carried out by minimizing the negative log likelihood $- \sum_n \log p(\Xb_f^{(n)} \mid \Xb_p^{(n)} ; \thetab)$, where $\thetab$ is a collection of all parameters from the encoder LSTM, the decoder LSTM, and the decoder outputs. This is carried out using stochastic gradient descent. We note that the encoder, the decoder, the decoder's outputs, and the negative log likelihood are all constructed within a single computation graph, and we can differentiate our loss with respect to all parameters automatically and efficiently using backpropagation through time \cite{werbos1990}. Our implementation is based on PyTorch.

\section{Experiments} \label{experiments}

Here we carry out two sets of experiments. First, we compare the predictions and encodings from our future-prediction model equipped with mixture density networks, which we refer to as FP MDN, with two baseline versions: FP -MDN, which focuses on future prediction without MDNs, and -FP MDN, which instead of predicting the future learns to compress and reconstruct the past in an autoencoder-like fashion. Second, we compare these approaches in an information-retrieval setting alongside the state-of-the-art approach \cite{gao2016ijcars}.

\subsection{Dataset}

The \anon{Minimally Invasive Surgical Training and Innovation Center - Science of Learning (MISTIC-SL)} dataset focuses on minimally-invasive, robot-assisted surgery using a \emph{da Vinci} surgical system, in which trainees perform a structured set of tasks (see Fig. \ref{query-vis}). We follow \cite{gao2016ijcars} and only use data from the 15 right-handed trainees in the study. Each trainee performed between 1 and 6 trials, for a total of 39 trials. We use 14 kinematic signals in all experiments: velocities, rotational velocities, and the gripper angle of the tooltip, all for both the left and right hands. In addition, experts manually annotated the trials so that all moments in time are associated with 1 of 4 high-level activities: Suture Throw (ST), Knot Tying (KT), Grasp Pull Run Suture (GPRS), or Intermaneuver Segment (IMS). We emphasize that these labels are not used in any way to obtain the encodings.

\subsection{Future Prediction}

\begin{table*}[t]
\centering
\scriptsize
\caption{Quantitative results for kinematics-based queries.}
\label{query-table}
\renewcommand{\arraystretch}{1.2}
\begin{tabularx}{\textwidth}{lYYY}
\toprule
                                      & Precision          & Recall             & F1 Score                  \\[1.25ex]
                                      & \multicolumn{3}{c}{\textbf{Suturing}} \\[1.25ex]
DAE + AS-DTW \cite{gao2016ijcars}     & 0.53 $\pm$ 0.15    & 0.75 $\pm$ 0.16    & 0.60 $\pm$ 0.14           \\[1.25ex]
-FP MDN                               & 0.50 $\pm$ 0.08    & 0.75 $\pm$ 0.07    & 0.59 $\pm$ 0.07           \\
FP -MDN                               & 0.54 $\pm$ 0.07    & 0.76 $\pm$ 0.08    & 0.62 $\pm$ 0.06           \\
FP MDN                                & 0.81 $\pm$ 0.06    & 0.74 $\pm$ 0.10    & \textbf{0.77 $\pm$ 0.05}  \\[1.25ex]
                                      & \multicolumn{3}{c}{\textbf{Knot Tying}} \\[1.25ex]
DAE + AS-DTW \cite{gao2016ijcars}     & ---                & ---                & ---           \\[1.25ex]
-FP MDN                               & 0.37 $\pm$ 0.05    & 0.73 $\pm$ 0.02    & 0.49 $\pm$ 0.05           \\
FP -MDN                               & 0.34 $\pm$ 0.05    & 0.74 $\pm$ 0.02    & 0.46 $\pm$ 0.05           \\
FP MDN                                & 0.62 $\pm$ 0.08    & 0.74 $\pm$ 0.04    & \textbf{0.67 $\pm$ 0.05}  \\[1.25ex]
\bottomrule
\end{tabularx}
\end{table*}

We train our model using 5 second windows of kinematics, extracted at random during training. Adam was used for optimization with a learning rate of 0.005, with other hyperparameters fixed to their defaults \cite{kingma2014}. We trained for 5000 steps using a batch size of 50 (approximately 50 epochs). The hyperparameters tuned in our experiments were $n_h$, the number of hidden units for the encoder and decoder LSTMs, and $n_c$, the number of mixture components. For hyperparameter selection, 4 subjects were held out for validation. We began overly simple with $n_h = 16$ and $n_c = 1$, and proceeded to double $n_h$ or $n_c$ whenever doing so improved the held-out likelihood. This led to final values of $n_h = 64$ and $n_c = 16$.

Results for the FP MDN and baselines are shown in Figure \ref{fp}, in which we show predictions, and in Figure \ref{tsne}, in which we show 2-D representations obtained with t-SNE \cite{maaten2008}. We can see that the addition of future prediction and MDNs leads to more separation between high-level activities in the encoding space.

\subsection{Information Retrieval with Motion-Based Queries}

Here we present results for retrieving kinematic frames based on a motion-based query, using the tasks of suturing and knot tying. We focus on the most difficult but most useful scenario: querying with a sequence from one subject $i$ and retrieving frames from other subjects $j \neq i$.

In order to retrieve kinematic frames, we form encodings using \emph{all windows} within one segment of an activity by subject $i$, compute the cosines between these encodings and all encodings for subject $j$, take the maximum (over windows) on a per-frame basis, and threshold. For evaluation, we follow \cite{gao2016ijcars}, computing each metric (precision, recall, and F1 score) from each source subject $i$ to each target subject $j \neq i$, and finally averaging over all target subjects.

Quantitative results are shown in Table \ref{query-table}, comparing the FP MDN to its baselines and the state-of-the-art approach \cite{gao2016ijcars}, and qualitative results are shown in Figure \ref{query-vis}. We can see that the FP MDN significantly outperforms the two simpler baselines, as well as the state-of-the-art approach in the case of suturing, improving from an F1 score of $0.60 \pm 0.14$ to $0.77 \pm 0.05$.

\section{Summary and Future Work} \label{summary}

We showed that it is possible to learn meaningful representations of surgical motion, without supervision, by searching for representations that can reliably predict the future. The usefulness of these representations was demonstrated in the context of information retrieval, where we used future prediction equipped with mixture density networks to improve the state-of-the-art performance for motion-based suturing queries from an F1 score of $0.60 \pm 0.14$ to $0.77 \pm 0.05$.

Because we do not rely on annotations, our method is applicable to arbitrarily large databases of surgical motion. From one perspective, exploring large databases using these encodings is exciting in and of itself. From another perspective, we also expect such encodings to improve downstream tasks such skill assessment and surgical activity recognition, especially in the regime of few annotations. Finally, as illustrated in Figure \ref{query-vis}, we believe that these encodings can also be used to aid the annotation process itself.\\

\noindent \textbf{Acknowledgements.} This work was supported by a fellowship for modeling, simulation, and training from the Link Foundation. We would also like to thank Anand Malpani, Swaroop Vedula, Gyusung I. Lee, and Mija R. Lee for procuring the MISTIC-SL dataset.

\bibliographystyle{splncs03}
\bibliography{miccai-2018}

\begin{thebibliography}{10}
\providecommand{\url}[1]{\texttt{#1}}
\providecommand{\urlprefix}{URL }

\bibitem{ahmidi2017}
Ahmidi, N., Tao, L., Sefati, S., Gao, Y., Lea, C., Haro, B.B., Zappella, L.,
  Khudanpur, S., Vidal, R., Hager, G.D.: A dataset and benchmarks for
  segmentation and recognition of gestures in robotic surgery. IEEE
  Transactions on Biomedical Engineering  64(9),  2025--2041 (2017)

\bibitem{bengio1994}
Bengio, Y., Simard, P., Frasconi, P.: Learning long-term dependencies with
  gradient descent is difficult. IEEE Transactions on Neural Networks  5(2),
  157--166 (1994)

\bibitem{bishop1994}
Bishop, C.M.: Mixture density networks. Tech. rep., Aston University (1994)

\bibitem{chen2016}
Chen, Z., Malpani, A., Chalasani, P., Deguet, A., Vedula, S.S., Kazanzides, P.,
  Taylor, R.H.: Virtual fixture assistance for needle passing and knot tying.
  In: Intelligent Robots and Systems (IROS). pp. 2343--2350 (2016)

\bibitem{cho2014}
Cho, K., van Merri{\"{e}}nboer, B., G{\"{u}}l{\c c}ehre, {\c C}., Bahdanau, D.,
  Bougares, F., Schwenk, H., Bengio, Y.: Learning phrase representations using
  {RNN} encoder--decoder for statistical machine translation. EMNLP  (2014)

\bibitem{despinoy2016}
Despinoy, F., Bouget, D., Forestier, G., Penet, C., Zemiti, N., Poignet, P.,
  Jannin, P.: Unsupervised trajectory segmentation for surgical gesture
  recognition in robotic training. IEEE Transactions on Biomedical Engineering
  63(6),  1280--1291 (2016)

\bibitem{dipietro2016}
DiPietro, R., Lea, C., Malpani, A., Ahmidi, N., Vedula, S.S., Lee, G.I., Lee,
  M.R., Hager, G.D.: Recognizing surgical activities with recurrent neural
  networks. International Conference on Medical Image Computing and
  Computer-Assisted Intervention pp. 551--558 (2016)

\bibitem{elman1990}
Elman, J.L.: Finding structure in time. Cognitive science  14(2),  179--211
  (1990)

\bibitem{gao2016ijcars}
Gao, Y., Vedula, S.S., Lee, G.I., Lee, M.R., Khudanpur, S., Hager, G.D.:
  Query-by-example surgical activity detection. International journal of
  computer assisted radiology and surgery  11(6),  987--996 (2016)

\bibitem{gao2016icra}
Gao, Y., Vedula, S., Lee, G.I., Lee, M.R., Khudanpur, S., Hager, G.D.:
  Unsupervised surgical data alignment with application to automatic activity
  annotation. 2016 IEEE International Conference on Robotics and Automation
  (ICRA)  (2016)

\bibitem{gers2000fg}
Gers, F.A., Schmidhuber, J., Cummins, F.: Learning to forget: Continual
  prediction with {LSTM}. Neural computation  12(10),  2451--2471 (2000)

\bibitem{hochreiter1997}
Hochreiter, S., Schmidhuber, J.: Long short-term memory. Neural computation
  9(8),  1735--1780 (1997)

\bibitem{kingma2014}
Kingma, D.P., Ba, J.: Adam: A method for stochastic optimization. arXiv
  preprint arXiv:1412.6980  (2014)

\bibitem{krishnan2017}
Krishnan, S., Garg, A., Patil, S., Lea, C., Hager, G., Abbeel, P., Goldberg,
  K.: Transition state clustering: Unsupervised surgical trajectory
  segmentation for robot learning. International Journal of Robotics Research
  36(13-14) (2017)

\bibitem{maaten2008}
Maaten, L.v.d., Hinton, G.: Visualizing data using {t-SNE}. Journal of machine
  learning research  9(Nov),  2579--2605 (2008)

\bibitem{reiley2008}
Reiley, C.E., Akinbiyi, T., Burschka, D., Chang, D.C., Okamura, A.M., Yuh,
  D.D.: Effects of visual force feedback on robot-assisted surgical task
  performance. The Journal of thoracic and cardiovascular surgery  135(1),
  196--202 (2008)

\bibitem{srivastava2015}
Srivastava, N., Mansimov, E., Salakhudinov, R.: Unsupervised learning of video
  representations using {LSTM}s. In: International conference on machine
  learning. pp. 843--852 (2015)

\bibitem{vedula2016}
Vedula, S.S., Malpani, A., Ahmidi, N., Khudanpur, S., Hager, G., Chen, C.C.G.:
  Task-level vs. segment-level quantitative metrics for surgical skill
  assessment. Journal of surgical education  73(3),  482--489 (2016)

\bibitem{werbos1990}
Werbos, P.J.: Backpropagation through time: what it does and how to do it.
  Proceedings of the IEEE  78(10),  1550--1560 (1990)

\bibitem{zia2017}
Zia, A., Zhang, C., Xiong, X., Jarc, A.M.: Temporal clustering of surgical
  activities in robot-assisted surgery. International journal of computer
  assisted radiology and surgery  12(7),  1171--1178 (2017)

\end{thebibliography}

\end{document}